# Decoding Market Emotions in Cryptocurrency Tweets via Predictive Statement Classification with Machine Learning and Transformers

Moein Shahiki Tash[a], Zahra Ahani[a], Mohim Tash[b], Mostafa Keikhay Farzaneh[b], Ari Y. Barrera-Animas[c], Olga Kolesnikova[a]

[a]*Centro de Investigación en Computación, Instituto Politécnico Nacional, , Ciudad de México, México*
[b]*Management and Economics Faculty, University of Sistan and Baluchestan, , Zahedan, Iran*
[c]*Universidad Panamericana, Augusto Rodin, Ciudad de México, México*

**Abstract**

The growing prominence of cryptocurrencies has triggered widespread public engagement and increased speculative activity, particularly on social media platforms. This study introduces a novel classification framework for identifying predictive statements in cryptocurrency-related tweets, focusing on five popular cryptocurrencies: Cardano, Matic, Binance, Ripple, and Fantom. The classification process is divided into two stages: Task 1 involves binary classification to distinguish between Predictive and Non-Predictive statements. Tweets identified as Predictive proceed to Task 2, where they are further categorized as Incremental, Decremental, or Neutral. To build a robust dataset, we combined manual and GPT-based annotation methods and utilized SenticNet to extract emotion features corresponding to each prediction category. To address class imbalance, GPT-generated paraphrasing was employed for data augmentation. We evaluated a wide range of machine learning, deep learning, and transformer-based models across both tasks. The results show that GPT-based balancing significantly enhanced model performance, with transformer models achieving the highest F1-score in Task 1, while traditional machine learning models performed best in Task 2. Furthermore, our emotion analysis revealed distinct emotional patterns associated with each prediction category across the different cryptocurrencies.

## 1. Introduction

From its early beginnings, the cryptocurrency market has experienced rapid growth, reaching a peak valuation of $3 trillion in November 2021. Cryptocurrencies are digital assets used for online transactions and can be exchanged for other cryptocurrencies or traditional fiat currencies. They operate independently of central authorities, with all transactions securely recorded on a blockchain (Pellicani et al., 2025).

This expanding market has increasingly attracted attention from a broad range of stakeholders, including investors, regulators, fund managers, policymakers, and researchers. This surge of interest followed the launch of the first cryptocurrency, Bitcoin (BTC), introduced in 2008 by the anonymous creator(s) known as Nakamoto (Nakamoto, 2008). Bitcoin's rise—from having no initial value in 2009 to reaching an all-time high of 103,900.47 USD on December 5, 2024—can be attributed to its pioneering features such as consensus mechanisms (Proof-of-Work and Proof-of-Stake) and secure ledger technology (Kehinde et al., 2025).

Parallel to this financial growth, the increasing prevalence of social media platforms has amplified public engagement with cryptocurrencies. Many users actively discuss these digital assets online, sharing their emotions and sentiments. Additionally, a growing number of individuals use social media data to predict cryptocurrency price movements, reflecting the community's desire to anticipate market trends (Dulău and Dulău, 2019).

Unlike traditional financial assets such as stocks or bonds, cryptocurrencies do not possess intrinsic value and are significantly influenced by speculative trading. This characteristic results in highly volatile and unpredictable price behavior, which in turn drives the need for robust forecasting and predictive models to help market participants navigate the crypto economy's complex dynamics (Islam et al., 2025; Sizan et al., 2023).

More generally, prediction is a broad concept involving the estimation of unknown or future outcomes based on available information and data, encompassing both the process and the resulting forecasts (Papacharalampous and Tyralis, 2022).

Investors continuously seek effective strategies to anticipate market trends and make informed decisions. However, the inherent volatility and unpredictability of cryptocurrency markets generate diverse emotional responses among participants. For instance, some may view a price decline as an attractive buying opportunity, feeling optimistic, while others may experience disappointment or frustration due to losses, influencing their perspectives and actions within the market. Emotion analysis on social media has become a valuable tool for investors to gauge the overall sentiment surrounding market prices (Tash et al., 2024b). However, it is crucial to distinguish between the emotions of users who view a price drop as a good investment opportunity and those who are expressing frustration or sadness due to financial loss. Making this distinction provides a more accurate understanding of price movements and the emotional landscape of investors.

A key motivation of our framework is that the same mar-ket movement can trigger opposite emotions in different users. When the price of a cryptocurrency increases, some investors express happiness because their holdings gained value, while others react negatively because they lost the opportunity to buy at a lower price. Conversely, when the price decreases, some participants express disappointment due to financial loss, while others react positively because they perceive it as a buying opportunity. Without separating these reactions, traditional sentiment analysis may provide misleading or ambiguous signals about investor expectations. Our multi-class categorization directly addresses this challenge by disentangling Incremental, Decremental, and Neutral predictive statements, allowing us to align emotions with directional market expectations. This distinction is essential for connecting social media discourse with decision-making processes in volatile cryptocurrency markets.

We introduce a new concept called predictive statement, which helps us distinguish between different types of user comments. This concept categorizes comments into four groups: Incremental predictions (indicating expected price increases), Decremental predictions (indicating expected price decreases), Neutral predictions (indicating no significant change), and Non-predictive statements (which do not involve any future expectation). By using this classification, we can more effectively separate users who are optimistic about market opportunities from those who are reacting emotionally to financial losses. We chose five widely recognized cryptocurrencies—Cardano, Matic, Binance, Ripple, and Fantom (Shahiki Tash et al., 2024)—and conducted both manual annotation and GPT-based labeling. After annotation, we utilized SenticNet to extract emotions including (Delight and Joy Enthusiasm and Eagerness Delight and Pleasantness Grief and Sadness Fear and Anxiety Rage and Anger) to identify whether comments were incremental, decremental, or neutral. Subsequently, we applied traditional machine learning, deep learning, and transformer models for classification.

The main contributions of this study can be summarized as follows:

- **Novel Categorization:** Introduced a new framework to classify crypto tweets into Incremental, Decremental, Neutral, and Non-Predictive statements.

- **Annotated Dataset:** Created a labeled dataset using both human and GPT-based annotation, with class balancing via GPT-generated paraphrases.

- **Emotion Analysis with SenticNet:** Applied SenticNet to extract and group emotions from crypto-related tweets, enabling analysis of emotional patterns linked to different cryptocurrencies.

- **Model Comparison:** Evaluated traditional ML, deep learning, and transformer models on binary and multi-class prediction tasks.

- **Effective Data Balancing:** Showed that GPT-based balancing significantly improves classification performance, especially in macro F1-score.

## 2. Definitions

### 2.1. Predictive Statement

In this task, a prediction refers to a statement about the future performance or trend of an investment or market within tweets. We categorize these predictions as incremental (indicating expected improvement), decremental (indicating expected decline), or neutral (indicating no significant change). Our goal is to analyze investment-related tweets to determine which category—incremental, decremental, or neutral—has the highest percentage of predictions.

- **Incremental predictions** refer to forecasts that indicate a positive trend or enhancement in a future event or outcome. These predictions suggest that there will be growth, improvement, or an upward shift in the situation being analyzed. For instance, in business, an incremental prediction might forecast a rise in sales or an increase in market share.

- **Decremental predictions** pertain to forecasts that signal a negative trend or decline in a future event or outcome. These predictions indicate a reduction, deterioration, or downward movement in the scenario under consideration. For example, in economics, a decremental prediction could suggest a drop in GDP or a decline in employment rates.

- **Neutral predictions** describe forecasts that predict no significant change in a future event or outcome. These predictions suggest stability or stasis, where the current conditions are expected to remain relatively unchanged. In environmental studies, a neutral prediction might imply that current pollution levels will stay constant over a given period.

- **Non-predictive** text segments are those that do not contain any forecasts or projections about future events or outcomes. These segments may provide descriptions, explanations, or analyses that are focused on past or present conditions without making any statements about what might happen in the future. Examples of different categories of Predictive statements can be found in Table 1.

| Criteria | Examples |
|---|---|
| **Incremental** | There will be a steady increase in market share over the next quarter. |
| **Incremental** | Profits are expected to double by the end of the year. |
| **Incremental** | The company anticipates growth in revenue due to new product launches. |
| **Decremental** | Sales are projected to decline in the next fiscal quarter. |
| **Decremental** | There will be a 20% decrease in production efficiency. |
| **Decremental** | The market forecasts a decrease in consumer confidence. |
| **Neutral** | The company expects revenue to remain consistent in the upcoming quarter. |
| **Neutral** | There is uncertainty regarding future market conditions. |
| **Non-Predictive** | Blockchain technology is revolutionizing various industries worldwide. |
| **Non-Predictive** | The company reported record profits for the current fiscal year. |

Table 1: Examples of Different Categories of Predictive Statements

## 3. Literature Review

Recent research in cryptocurrency forecasting has primarily focused on Bitcoin and often relies on technical indicators and limited input features. In contrast, (Viéitez et al., 2024) proposed a machine learning-based framework specifically for Ethereum price and trend prediction using only contextual features such as traditional stock indices, online statistics, and sentiment analysis from Reddit and Google News, excluding technical indicators entirely. They implemented GRU and LSTM networks for price regression and support vector machines for trend classification, applying multiple feature selection methods to optimize input variables. Importantly, the authors validated their models through the design of two novel knowledge-based investment strategies, demonstrating practical profitability in different market conditions. This integrated approach—combining predictive modeling, sentiment analysis, feature selection, and real-market strategy evaluation—offers a more comprehensive and generalizable methodology compared to prior work, which often lacks full-cycle validation or focuses narrowly on Bitcoin.

(Qureshi et al., 2025) – Evaluating ML Models for Predictive Accuracy in Cryptocurrency Price Forecasting Qureshi present a comprehensive evaluation of machine learning models for algorithmic trading in the volatile cryptocurrency market. Their focus lies in comparing classification models such as random forest, logistic regression, and gradient boosting using historical price data, emphasizing technical indicators and hyperparameter tuning. They also address class imbalance using resampling techniques like SMOTE and highlight the significance of backtesting for assessing trading strategies in real-world scenarios. Relation to our work: Unlike Qureshi et al., who focus on forecasting price movements using technical features, our work emphasizes predictive sentiment analysis by categorizing social media content (e.g., tweets) into Incremental, Decremental, Neutral, and Non-Predictive categories. While their study is grounded in numerical market indicators, our model incorporates language-based investor cues, offering a complementary perspective on market forecasting.

(Kumar et al., 2024) – Stock and Cryptocurrency Price Prediction using SVM and LSTM Kumar propose a dual-model approach using LSTM for time-series analysis and SVM for classification to predict future stock and cryptocurrency prices. The system leverages live market data fetched using the yFinance library, with predictions visualized through a Streamlit-based dashboard. Their model reports high accuracy (above 95%) and integrates multiple ML techniques like decision trees and random forests. Relation to our work: While Kumar et al. focus on direct price prediction from historical numerical data, our research shifts the focus toward linguistic modeling by identifying predictive language patterns in social media posts. Rather than predicting raw prices, we classify sentiment-laden investor statements, offering a new layer of interpretability and potential integration with price-based models for hybrid decision systems.

(Golnari et al., 2024) proposed a probabilistic deep learning framework for cryptocurrency price prediction, introducing a P-GRU model that outputs predictive distributions rather than point estimates. Their approach incorporates Bayesian neural networks to quantify uncertainty, alongside a customized R2-based callback mechanism to optimize model weights during training. Furthermore, they implemented a transfer learning strategy by pretraining the model on Bitcoin data and adapting it to six other cryptocurrencies, achieving superior performance across all assets. This combination of probabilistic forecasting and transfer learning enhances both model robustness and generalizability under volatile market conditions.

(Otabek and Choi, 2024) conducted a comprehensive review examining the relationship between cryptocurrency price prediction models and their real-world impact on trading strategies. Unlike prior works that treat forecasting and trading as separate areas, their study integrates econometric, statistical, machine learning, and sentiment analysis models to assess how predictive accuracy directly enhances trading performance. They also introduce a taxonomy of techniques and evaluation metrics while highlighting recent advances in AI-based trading strategies, including reinforcement learning and algorithmic systems. By focusing on BTC, ETH, and LTC, the authors emphasize practical profitability and adaptivity in volatile markets, making their work a crucial reference point for understanding the intersection of predictive modeling and trading efficacy.

In summary, prior studies have either concentrated on numerical market indicators for price prediction or on coarse-grained sentiment analysis (positive vs. negative). None, to the best of our knowledge, have explicitly separated the *directional predictive intent* of user statements (Incremental, Decremental, Neutral) from their emotional tone. This distinction is essential because the same price movement can produce opposite emotions depending on the investor's perspective, and ignoring this leads to ambiguous forecasting signals. Our work fills this gap by introducing predictive statement classification as a novel intermediate layer between sentiment and price, thereby complementing existing numerical and sentiment-based approaches and opening new avenues for decision-support in cryptocurrency markets.

## 4. Methodolghy

In this study, we examined five popular cryptocurrencies: Cardano, Binance, Matic, Ripple, and Fantom. Our approach involved annotating the data through both manual and automated methods, introducing a new concept within the cryptocurrency field called predictive statements. These statements were classified into four categories: Incremental predictions, Decremental predictions, Neutral predictions, and Non-predictive statements. Additionally, we employed SenticNet to extract emotions such as Delight and Joy, Enthusiasm and Eagerness, Delight and Pleasantness, Grief and Sadness, Fear and Anxiety, and Rage and Anger, aiming to determine whether comments expressed incremental, decremental, or neutral sentiments.

To address class imbalance in the dataset, we used a GPT-based paraphrasing approach to generate synthetic variations of the original data. Following data preparation, we applied a range of machine learning algorithms to perform the prediction task. Finally, we conducted a comparative analysis of the performance of the different models.

*4.1. Data Collection*

The data for our analysis was obtained from (Tash et al., 2024a), who collected English tweets from the X platform covering the period from September 2021 to March 2023. The original dataset comprised approximately 30,000 tweets, from which we randomly selected 3,116 tweets related to the five cryptocurrencies (Cardano, Binance, Fantom, Matic, and Ripple) for annotation.

*4.2. Annotator selection*

After finalizing our dataset, we recruited two male annotators with strong proficiency in English, aged between 28 and 30, both holding a Master's degree in Computer Science. To assess their annotation capabilities, we initially provided them with a set of 100 sample instances. Based on their performance, we reviewed their outputs, identified any issues, and provided feedback to improve consistency and accuracy. Following this calibration phase, the annotators proceeded with the full annotation task over the course of one month. Additionally, we employed GPT models as a third annotator to complement the human annotations and enrich the overall labeling process.

*4.3. Annotation guidelines*

This annotation guideline is designed to help annotators consistently identify and classify predictive statements in cryptocurrency-related tweets. The goal is to determine whether a tweet expresses a forecast about the future performance or trend of a cryptocurrency or market, and if so, to assign it to one of the defined categories: Incremental, Decremental, Neutral, or Non-predictive.

*4.3.1. Objective*

Annotators will examine each tweet and determine whether it contains a prediction about the future. If a prediction is present, they must classify it as Incremental, Decremental, or Neutral. If no prediction is found, the tweet should be labeled as Non-predictive.

*4.3.2. Definitions and Labeling Criteria*

**A. Incremental Prediction (Label: 1)** Indicates a positive future outcome or improvement.

Common cues: will increase, going to rise, bullish, expect growth, price is headed up, will break ATH, etc.

Example: "Cardano is going to hit $1.5 soon, the momentum is strong!"

**B. Decremental Prediction (Label: 2)** Indicates a negative future outcome or decline.

Common cues: will drop, likely to fall, bearish, expect loss, dump incoming, etc.

Example: "Bitcoin is losing support fast. I think we're headed below $20K."

**C. Neutral Prediction (Label: 3)** Predicts no significant change in the near future.

Common cues: will remain stable, no major movement expected, sideways trading likely, etc.

Example: "MATIC seems to be consolidating. I don't expect much price action this week."

**D. Non-Predictive (Label: 0)** Does not include any forecast about the future.

Focuses on past or current events, general opinions, or factual information.

Example: "Ethereum has a strong community and innovative developers."

*4.3.3. General Annotation Rules*

Focus only on statements that imply a future outcome.

If multiple predictions exist in one tweet, choose the dominant prediction (i.e., the one most emphasized or relevant).

Ignore sarcastic or unclear predictions unless they explicitly mention future outcomes.

Do not label speculative questions (e.g., "Will Bitcoin rise?") as predictive unless the author clearly implies an answer.

*4.3.4. Data Evaluation*

Inter-annotator agreement (IAA) serves as a metric to assess the level of consistency between annotators. In our evaluation, we measured the performance of the GPT-4o model by comparing its annotations to those made manually, using Cohen's Kappa Coefficient. The model achieved a score of 0.7493 for predictive statement detection, demonstrating strong alignment with human annotations. This result underscores the reliability of our dataset and the rigor of the annotation process.

*4.4. Statistics of the dataset*

Table 2 presents the distribution of labels across Task 1 and Task 2. In Task 1, the dataset consists of 2,000 Non-Predictive and 1,116 Predictive instances. Task 2 further categorizes the Predictive instances into three subtypes: 570 Incremental, 434 Decremental, and 112 Neutral predictions. This structured labeling allows for a more fine-grained analysis of predictive language in cryptocurrency-related content.

Table 2: Label Distribution for Task 1 and Task 2

| Label Category | Count |
|---|---|
| **Task 1** | |
| Non-Predictive | 2000 |
| Predictive | 1116 |
| **Total Task 1** | **3116** |
| **Task 2 (Subcategories of Predictive)** | |
| Predictive Incremental | 570 |
| Predictive Decremental | 434 |
| Predictive Neutral | 112 |
| **Total Task 2** | **1116** |

*4.5. Data Preprocessing*

After acquiring the dataset, we initiated a multi-step data preprocessing protocol aimed at refining and optimizing the data. This process comprised the following primary stages:

**URL Removal**: Employing a regular expression pattern, we systematically identified and eliminated any URLs present within the dataset.

**Text Cleaning**: This phase involved the systematic removal of special characters, such as punctuation marks, leveraging a dedicated dictionary of such characters. Additionally, words with a length equal to or less than two characters were excluded. The outcome was a refined version of the textual data devoid of unnecessary elements.

For each comment, we implemented preprocessing procedures to ensure its compatibility for subsequent analysis, encompassing:

- Elimination of superfluous characters or noise.
- Tokenization of comments to facilitate improved handling by the model.
- Normalization of text to ensure uniformity and consistency throughout the dataset.

## 5. Balanced data set

*5.1. GPT*

To balance our datasets for Task 1 and Task 2, we used OpenAI's GPT models to generate paraphrased versions of underrepresented classes. First, we loaded the original annotated dataset and identified class imbalances in both tasks. Then, for each minority class, we sampled examples and prompted GPT-4-o to produce paraphrases that preserved the original meaning but varied the wording. These synthetic paraphrased instances were added to the dataset to equalize class distributions. The process included a delay to respect API rate limits, and the final balanced datasets were saved for subsequent model training and evaluation.

*5.2. statistical data*

The balanced dataset consists of 2,000 instances each for the Predictive and Non-Predictive classes in Task 1. Within the Predictive class, Task 2 divides the data nearly equally into three subcategories: Neutral, Incremental, and Decremental predictions. Such a balanced distribution helps prevent model bias toward any class, facilitating more reliable and fair evaluation (see Table 3).

## 6. Model Training and Evaluation

Prior to model training, the dataset was preprocessed through several steps, including text cleaning, tokenization, and normalization, to optimize model performance. To ensure reliable evaluation, we implemented 5-fold cross-validation, providing a comprehensive assessment across both traditional machine learning methods and advanced deep learning approaches.

Table 3: Label Distribution for Task 1 and Task 2 (Balanced Dataset)

| Label Category | Count |
|---|---|
| **Task 1** | |
| Predictive | 2000 |
| Non-Predictive | 2000 |
| **Total (Task 1)** | **4000** |
| **Task 2 (Subcategories of Predictive)** | |
| Predictive Neutral | 570 |
| Predictive Incremental | 570 |
| Predictive Decremental | 570 |
| **Total (Task 2)** | **1710** |

For traditional models, we utilized standard feature extraction techniques such as TF-IDF, while advanced models incorporated deep learning architectures and transformer-based pre-trained models to capture richer semantic representations. We evaluated model performance using accuracy, precision, recall, and F1-score, reported with both macro and weighted averages.

The macro F1-score, which gives equal importance to each class by averaging their individual F1-scores, is particularly appropriate for balanced datasets like ours. In contrast, the weighted F1-score takes class distribution into account by assigning more weight to larger classes. Despite the availability of both metrics, we selected the macro F1-score as our primary evaluation criterion due to its balanced treatment of all classes.

The equations used to calculate Accuracy, Precision, Recall, and F1-score are adopted from (Derczynski, 2016).

**Accuracy** is defined as:

$$\text{Accuracy} = \frac{TP + TN}{TP + TN + FP + FN} \quad (1)$$

**Precision**, **Recall**, and **F1-score** are defined as:

$$\text{Precision} = \frac{TP}{TP + FP} \quad (2)$$

$$\text{Recall} = \frac{TP}{TP + FN} \quad (3)$$

$$\text{F1-score} = 2 \times \frac{\text{Precision} \times \text{Recall}}{\text{Precision} + \text{Recall}} \quad (4)$$

**Macro F1-score** calculates the F1-score for each class separately and then averages them:

$$\text{Macro-F1} = \frac{1}{N} \sum_{i=1}^{N} \text{F1-score}_i \quad (5)$$

where *N* is the number of classes.

**Weighted F1-score** takes class imbalance into account by weighting each class's F1-score based on its support:

$$\text{Weighted-F1} = \sum_{i=1}^{N} \frac{\text{support}_i}{\text{total instances}} \times \text{F1-score}_i \quad (6)$$

**Weighted Precision and Recall** are computed analogously:

$$\text{Weighted-Precision} = \sum_{i=1}^{\aleph} \frac{\text{support}_i}{\text{total instances}} \times \text{Precision}_i \quad (7)$$

$$\text{Weighted-Recall} = \sum_{i=1}^{\aleph} \frac{\text{support}_i}{\text{total instances}} \times \text{Recall}_i \quad (8)$$

where $\text{support}_i$ is the number of instances belonging to class $i$.

*6.1. Emotion*

We employed SenticNet (SenticNet), a sentiment analysis tool, to extract emotional insights for each cryptocurrency, as outlined in Table 4 with supporting explanations. Notably, the emotional percentages associated with each tweet were derived from SenticNet's output. To enhance our analysis, we grouped similar emotions—such as Fear and Anxiety, Rage and Anger, Grief and Sadness, Delight and Pleasantness, Enthusiasm and Eagerness, and Delight and Joy—and computed their average values across the five cryptocurrencies and their respective predictive categories.

*6.2. Overview of Models for Text Classification*

This study evaluated a range of machine learning methods for cryptocurrency-related text classification, spanning traditional, deep learning, and transformer-based approaches.

*Traditional Machine Learning.* We employed Logistic Regression, Support Vector Machines (both linear and RBF kernels), XGBoost, and Random Forest due to their effectiveness in text classification tasks (Ahani et al., 2024). TF-IDF vectorization was used to convert text into numerical features, capturing term importance relative to the dataset (Roelleke and Wang, 2008).

*Deep Learning.* Deep neural networks such as CNN and BiLSTM were utilized, each combined with pre-trained word embeddings GloVe and FastText. These embeddings help the models capture semantic and contextual nuances in the data, improving predictive accuracy (Kolesnikova et al., 2025).

*Transformers.* We leveraged several pre-trained transformer models from the Hugging Face library, including DistilRoBERTa-base, RoBERTa-base, RoBERTuito, BERT-base-uncased, and XLM-RoBERTa (papluca/xlm-roberta-base-language-detection). These models, powered by self-attention mechanisms, demonstrated strong capabilities in understanding complex language patterns and achieving superior classification results (Kolesnikova et al., 2025).

**7. Results and Analysis**

*7.1. Emotion*

Table 4 shows the distribution of emotion categories across cryptocurrencies and predictive labels. Across all cryptocurrencies, Delight and Joy consistently emerges as the most dominant emotional category, especially in the Incremental class—most notably for FTM (81.82), MATIC (72.41), and ADA (61.29). This suggests that users expressing optimism about future market increases tend to use highly positive emotional tones. Enthusiasm and Eagerness also aligns with upward market expectations, peaking in MATIC (49.66) and ADA (44.09). On the other hand, Grief and Sadness, Fear and Anxiety, and Rage and Anger appear more frequently in Decremental comments, particularly for BNB, ADA, and XRP, indicating emotionally negative reactions to expected downturns. Interestingly, Neutral predictions still exhibit substantial Delight and Joy and Delight and Pleasantness, suggesting that even when users are not making strong directional forecasts, emotional positivity remains present—potentially due to general optimism or long-term confidence in the market. Note: values represent the proportion of tweets in a given predictive category that were associated with each emotion, as identified by SenticNet. Percentages in each row do not add up to 100 because multiple emotions can co-occur in a single tweet. A double dash (–) indicates a zero value.

*7.2. Traditional machine learning*

This section presents the performance of traditional machine learning models on both Task 1 (Predictive vs. Non-Predictive) and Task 2 (Incremental, Decremental, Neutral) using unbalanced and balanced datasets. Evaluation metrics include precision, recall, and F1-score, reported with both weighted and macro averages. Accuracy is also provided for completeness. The macro F1-score was selected as the primary metric for comparison, given the importance of treating each class equally in balanced classification scenarios. For Tables 5 to 12, the reported metrics follow the same format: weighted (W) and macro (M) averages for Precision (Prec.), Recall (Rec.), and F1-score (F1), along with Accuracy (Acc.).

*7.2.1. Results on the Original (Unbalanced) Dataset*

Table 5 and Table 6 report the performance of five traditional classifiers on the original, unbalanced dataset.

In Task 1, the best performance in terms of macro F1-score was achieved by the XGBoost model (0.5068), followed closely by the SVM with a linear kernel (0.5044).

In Task 2, all models demonstrated significantly lower macro F1-scores, with XGBoost performing best (0.3631). The macro F1-scores across all models in Task 2 highlight the challenge posed by class imbalance when trying to predict fine-grained categories (incremental, decremental, and neutral predictions).

*7.2.2. Results on the Balanced Dataset*

After balancing the dataset using GPT-generated paraphrases, the models were re-evaluated (see Table 7 and Table 8).

Table 4: Percentage of Emotion Categories per Cryptocurrency and Task 2 Label for Predictive Comments.

| Cryptocurrency | Task 2 Label | Delight and Joy | Enthusiasm and Eagerness | Delight and Pleasantness | Grief and Sadness | Fear and Anxiety | Rage and Anger |
|---|---|---|---|---|---|---|---|
| BNB | Neutral | 36.84 | 15.79 | 36.84 | 5.26 | 31.58 | 5.26 |
| | Decremental | 44.09 | 26.88 | 16.13 | 24.73 | 13.98 | 10.75 |
| | Incremental | 47.57 | 40.78 | 30.10 | 12.62 | 20.39 | 2.91 |
| MATIC | Decremental | 58.33 | 42.86 | 45.24 | 7.14 | 2.38 | 2.38 |
| | Incremental | 72.41 | 49.66 | 39.31 | 5.52 | 1.38 | 0.69 |
| | Neutral | 68.18 | 63.64 | 31.82 | 4.55 | 4.55 | 4.55 |
| ADA | Decremental | 59.52 | 42.86 | 25.00 | 13.10 | 7.14 | 16.67 |
| | Incremental | 61.29 | 44.09 | 37.63 | 15.05 | 4.30 | 8.60 |
| | Neutral | 46.15 | 34.62 | 30.77 | 19.23 | 3.85 | – |
| FTM | Decremental | 66.67 | 26.39 | 20.83 | 16.67 | 5.56 | 6.94 |
| | Incremental | 81.82 | 41.82 | 26.36 | 6.36 | 3.64 | 2.73 |
| | Neutral | 83.33 | 41.67 | 37.50 | 12.50 | 8.33 | 4.17 |
| XRP | Incremental | 58.33 | 32.41 | 32.41 | 9.26 | 1.85 | 6.48 |
| | Decremental | 52.69 | 27.96 | 32.26 | 16.13 | 5.38 | 11.83 |
| | Neutral | 57.14 | 28.57 | 52.38 | 9.52 | – | 14.29 |

Table 5: Performance of Traditional ML Models on Task 1 (Unbalanced Dataset)

| Model | W-Prec | W-Rec | W-F1 | M-Prec | M-Rec | M-F1 | Accuracy |
|---|---|---|---|---|---|---|---|
| Logistic Regression | 0.6050 | 0.6435 | 0.5750 | 0.5835 | 0.5339 | 0.4986 | 0.6435 |
| Random Forest | 0.5246 | 0.5725 | 0.5360 | 0.4811 | 0.4868 | 0.4697 | 0.5725 |
| XGBoost | 0.5600 | 0.5969 | 0.5666 | 0.5236 | 0.5173 | 0.5068 | 0.5969 |
| SVM Linear | 0.5791 | 0.6239 | 0.5734 | 0.5484 | 0.5270 | 0.5044 | 0.6239 |
| SVM RBF | 0.5651 | 0.6200 | 0.5579 | 0.5315 | 0.5151 | 0.4821 | 0.6200 |

Table 6: Performance of Traditional ML Models on Task 2 (Unbalanced Dataset)

| Model | W-Prec | W-Rec | W-F1 | M-Prec | M-Rec | M-F1 | Accuracy |
|---|---|---|---|---|---|---|---|
| Logistic Regression | 0.4641 | 0.5269 | 0.4791 | 0.3409 | 0.3699 | 0.3436 | 0.5269 |
| Random Forest | 0.4560 | 0.5009 | 0.4467 | 0.3809 | 0.3546 | 0.3313 | 0.5009 |
| XGBoost | 0.4711 | 0.5063 | 0.4799 | 0.3779 | 0.3713 | 0.3631 | 0.5063 |
| SVM Linear | 0.4705 | 0.5323 | 0.4860 | 0.3461 | 0.3747 | 0.3493 | 0.5323 |
| SVM RBF | 0.4897 | 0.5323 | 0.4361 | 0.3646 | 0.3593 | 0.3034 | 0.5323 |

The results show a substantial improvement across all evaluation metrics.

For Task 1, the best macro F1-score was achieved by the SVM with RBF kernel (0.6825), followed closely by Logistic Regression (0.6672) and SVM Linear (0.6583). The improved performance across all models indicates that balancing helped mitigate bias toward the majority class.

In Task 2, macro F1-scores increased markedly compared to the unbalanced version. SVM with RBF kernel again performed best (0.6478), followed by Random Forest (0.6488) and SVM Linear (0.6463). This improvement confirms that the augmentation strategy effectively supported the model's ability to distinguish between the three types of predictions.

Table 7: Performance of Traditional ML Models on Task 1 (Balanced Dataset)

| Model | W-Prec | W-Rec | W-F1 | M-Prec | M-Rec | M-F1 | Accuracy |
|---|---|---|---|---|---|---|---|
| Logistic Regression | 0.6704 | 0.6683 | 0.6672 | 0.6704 | 0.6683 | 0.6672 | 0.6683 |
| Random Forest | 0.6431 | 0.6400 | 0.6380 | 0.6431 | 0.6400 | 0.6380 | 0.6400 |
| XGBoost | 0.6470 | 0.6420 | 0.6389 | 0.6470 | 0.6420 | 0.6389 | 0.6420 |
| SVM Linear | 0.6648 | 0.6605 | 0.6583 | 0.6648 | 0.6605 | 0.6583 | 0.6605 |
| SVM RBF | 0.7039 | 0.6885 | 0.6825 | 0.7039 | 0.6885 | 0.6825 | 0.6885 |

### 7.2.3. Comparison and Analysis

Across both tasks, balancing the dataset led to consistent improvements in macro-level metrics. This suggests that the models were better able to learn distinguishing features of the minority classes once more representative training examples were introduced. Weighted metrics also improved, though to a lesser extent, which reflects the shift toward a more even label distribution.

These results demonstrate the value of data augmentation in classification tasks with inherent label imbalance, particularly when using macro-averaged F1 as a primary evaluation metric.

### 7.3. Deep learning

To evaluate the performance of Deep Learning models on our classification tasks, we employed Convolutional Neural Networks (CNN) and Bi-directional Long Short-Term Memory (BiLSTM) models using GloVe and FastText word embeddings. The models were assessed on both unbalanced and balanced datasets for Task 1 (Predictive vs. Non-Predictive) and Task 2 (Predictive: Incremental, Decremental, Neutral).

#### 7.3.1. Task 1 – Unbalanced Dataset

Table 9 presents the performance of the models on the unbalanced Task 1 dataset. The BiLSTM model using GloVe embeddings outperformed other models with a weighted F1-score of 0.5577 and macro F1-score of 0.5070. In general, BiLSTM models slightly outperformed CNNs, indicating their ability to better capture contextual dependencies in longer texts. However, macro F1-scores remained relatively low (under 0.51), reflecting the imbalance in the dataset that penalized performance on the minority class.

#### 7.3.2. Task 2 – Unbalanced Dataset

As seen in Table 10, all models struggled with the more fine-grained three-class classification of predictive statements. The best macro F1-score achieved was 0.3656 by the CNN (FastText), showing that models had difficulty distinguishing between Incremental, Decremental, and Neutral statements due to both class imbalance and high semantic similarity among categories.

Table 8: Performance of Traditional ML Models on Task 2 (Balanced Dataset)

| Model | W-Prec | W-Rec | W-F1 | M-Prec | M-Rec | M-F1 | Accuracy |
|---|---|---|---|---|---|---|---|
| Logistic Regression | 0.6396 | 0.6413 | 0.6397 | 0.6395 | 0.6412 | 0.6396 | 0.6413 |
| Random Forest | 0.6503 | 0.6513 | 0.6489 | 0.6503 | 0.6512 | 0.6488 | 0.6513 |
| XGBoost | 0.6075 | 0.6068 | 0.6060 | 0.6075 | 0.6067 | 0.6059 | 0.6068 |
| SVM Linear | 0.6480 | 0.6454 | 0.6463 | 0.6479 | 0.6453 | 0.6463 | 0.6454 |
| SVM RBF | 0.6625 | 0.6524 | 0.6479 | 0.6624 | 0.6523 | 0.6478 | 0.6524 |

Table 9: Performance of Deep Learning Models on Task 1 (Unbalanced Dataset)

| Model | W-Prec | W-Rec | W-F1 | M-Prec | M-Rec | M-F1 | Accuracy |
|---|---|---|---|---|---|---|---|
| CNN (GloVe) | 0.5424 | 0.5577 | 0.5467 | 0.5022 | 0.5028 | 0.4986 | 0.5577 |
| BiLSTM (GloVe) | 0.5514 | 0.5728 | 0.5577 | 0.5125 | 0.5116 | 0.5070 | 0.5728 |
| CNN (FastText) | 0.5254 | 0.5430 | 0.5320 | 0.4833 | 0.4846 | 0.4816 | 0.5430 |
| BiLSTM (FastText) | 0.5288 | 0.5555 | 0.5369 | 0.4866 | 0.4894 | 0.4821 | 0.5555 |

Table 10: Performance of Deep Learning Models on Task 2 (Unbalanced Dataset)

| Model | W-Prec | W-Rec | W-F1 | M-Prec | M-Rec | M-F1 | Accuracy |
|---|---|---|---|---|---|---|---|
| CNN (GloVe) | 0.4675 | 0.5008 | 0.4778 | 0.3778 | 0.3699 | 0.3615 | 0.5008 |
| BiLSTM (GloVe) | 0.4732 | 0.5026 | 0.4790 | 0.3766 | 0.3723 | 0.3630 | 0.5026 |
| CNN (FastText) | 0.4655 | 0.4954 | 0.4748 | 0.3756 | 0.3723 | 0.3656 | 0.4954 |
| BiLSTM (FastText) | 0.4500 | 0.4695 | 0.4560 | 0.3497 | 0.3568 | 0.3485 | 0.4695 |

*7.3.3. Task 1 – Balanced Dataset*

After dataset balancing using GPT-based paraphrasing, model performance improved significantly as shown in Table 11. All models achieved macro F1-scores above 0.62. The BiLSTM (GloVe) reached the highest macro F1-score of 0.6349, demonstrating that deep learning models can effectively learn class boundaries when class imbalance is mitigated. This improvement is consistent across both precision and recall.

*7.4. Task 2 – Balanced Dataset*

The balanced dataset also positively impacted Task 2 results (Table 12). The CNN model with GloVe embeddings achieved the best macro F1-score of 0.6149. While the improvement was not as dramatic as in Task 1, it highlights that the use of paraphrased data helped deep learning models better generalize across all three predictive categories. The BiLSTM (FastText) model followed closely with a macro F1-score of 0.6014.

*7.4.1. Comparison with Traditional Models*

Compared to traditional machine learning models, deep learning approaches—especially BiLSTM and CNN with GloVe—offered competitive or better performance, particularly on the balanced datasets. While traditional models such as SVM with RBF kernel performed well on Task 1, they lagged in Task 2 performance. This suggests that deep learning models are better suited for nuanced sentiment and intent prediction tasks such as those in Task 2.

Table 11: Performance of Deep Learning Models on Task 1 (Balanced Dataset)

| Model | W-Prec | W-Rec | W-F1 | M-Prec | M-Rec | M-F1 | Accuracy |
|---|---|---|---|---|---|---|---|
| CNN (GloVe) | 0.6671 | 0.6430 | 0.6277 | 0.6671 | 0.6430 | 0.6277 | 0.6430 |
| BiLSTM (GloVe) | 0.6761 | 0.6490 | 0.6349 | 0.6761 | 0.6490 | 0.6349 | 0.6490 |
| CNN (FastText) | 0.6619 | 0.6425 | 0.6231 | 0.6619 | 0.6425 | 0.6231 | 0.6425 |
| BiLSTM (FastText) | 0.6353 | 0.6353 | 0.6222 | 0.6353 | 0.6353 | 0.6222 | 0.6353 |

*7.5. Transformers*

Transformer-based models, including various configurations of RoBERTa, BERT, and XLM-RoBERTa, were employed to assess their performance in detecting predictive statements and their types in cryptocurrency-related tweets.

Table 12: Performance of Deep Learning Models on Task 2 (Balanced Dataset)

| Model | W-Prec | W-Rec | W-F1 | M-Prec | M-Rec | M-F1 | Accuracy |
|---|---|---|---|---|---|---|---|
| CNN (GloVe) | 0.6750 | 0.6336 | 0.6149 | 0.6750 | 0.6336 | 0.6149 | 0.6336 |
| BiLSTM (GloVe) | 0.6189 | 0.5886 | 0.5738 | 0.6187 | 0.5885 | 0.5737 | 0.5886 |
| CNN (FastText) | 0.6601 | 0.6254 | 0.6154 | 0.6600 | 0.6253 | 0.6153 | 0.6254 |
| BiLSTM (FastText) | 0.6329 | 0.6120 | 0.6016 | 0.6327 | 0.6118 | 0.6014 | 0.6120 |

*7.5.1. Task 1 – Unbalanced Dataset*

Table 13 illustrates the performance of the models on Task 1. Among the configurations, the roberta-base model achieved the highest weighted F1-score (0.6031) and macro F1-score (0.5578). The distilroberta-base variant also showed competitive performance, slightly outperforming BERT. Interestingly, xlm-roberta-base, a multilingual model, achieved the highest accuracy (0.6486) but had a lower macro F1-score (0.4301), indicating potential bias toward the dominant class in the unbalanced data. These results show that while transformers capture contextual meaning effectively, data imbalance remains a challenge, especially for minority classes.

*7.5.2. Task 2 – Unbalanced Dataset*

For the more complex multi-class Task 2, all transformer models demonstrated reduced macro F1-scores due to the increased difficulty of distinguishing between Incremental, Decremental, and Neutral predictions (Table 14). The best-performing model, distilroberta-base, only achieved a macro F1-score of 0.4051, further reinforcing the impact of class imbalance on fine-grained classification tasks.

Table 13: Transformer Model Results for Task 1 (Unbalanced Dataset)

| Model | $P_w$ | $R_w$ | $F1_w$ | $P_m$ | $R_m$ | $F1_m$ | Accuracy |
|---|---|---|---|---|---|---|---|
| DistilRoBERTa | 0.5842 | 0.5960 | 0.5882 | 0.5492 | 0.5457 | 0.5453 | 0.5960 |
| RoBERTa-base | 0.5988 | 0.6159 | 0.6031 | 0.5670 | 0.5584 | 0.5578 | 0.6159 |
| Robertuito | 0.5735 | 0.5818 | 0.5764 | 0.5367 | 0.5352 | 0.5345 | 0.5818 |
| BERT-base | 0.5634 | 0.5812 | 0.5692 | 0.5265 | 0.5230 | 0.5213 | 0.5812 |
| XLM-RoBERTa | 0.4611 | 0.6486 | 0.5297 | 0.3855 | 0.5183 | 0.4301 | 0.6486 |

Table 14: Transformer Model Results for Task 2 (Unbalanced Dataset)

| Model | $P_w$ | $R_w$ | $F1_w$ | $P_m$ | $R_m$ | $F1_m$ | Accuracy |
|---|---|---|---|---|---|---|---|
| DistilRoBERTa | 0.5175 | 0.5350 | 0.5215 | 0.4145 | 0.4078 | 0.4051 | 0.5350 |
| RoBERTa-base | 0.4136 | 0.5278 | 0.4522 | 0.3106 | 0.3783 | 0.3294 | 0.5278 |
| Robertuito | 0.4859 | 0.5090 | 0.4911 | 0.3938 | 0.3862 | 0.3817 | 0.5090 |
| BERT-base | 0.4816 | 0.4937 | 0.4862 | 0.3780 | 0.3793 | 0.3759 | 0.4937 |
| XLM-RoBERTa | 0.4400 | 0.5260 | 0.4681 | 0.3440 | 0.3753 | 0.3396 | 0.5260 |

*7.6. Task 1 – Balanced Dataset*

Table 15 shows a substantial improvement in performance after balancing the dataset. The xlm-roberta-base model attained the highest macro F1-score (0.7011), followed by roberta-base (0.6625). All models improved their precision and recall scores, confirming that addressing class imbalance allows transformers to generalize more effectively across all classes.

*7.7. Task 2 – Balanced Dataset*

Table 16 presents results on the balanced Task 2 dataset. While performance was still lower than in Task 1, all models demonstrated significant gains compared to the unbalanced setup. The distilroberta-base achieved a macro F1-score of 0.5936, outperforming other configurations. This suggests that transformer models, when trained on well-distributed data, can effectively capture nuanced predictive sentiment types, even in a fine-grained classification setting.

*7.8. Comparison with Traditional and Deep Learning Models*

Transformer models clearly outperform traditional ML and deep learning approaches on both tasks when the data is balanced. In particular, the contextual representations learned by transformers enable better generalization across complex semantic classes. Their superior macro F1-scores indicate a more equitable treatment of each class, validating their suitability for social media text classification tasks involving nuanced financial sentiment.

Table 15: Transformer Model Results for Task 1 (Balanced Dataset)

| Model | $P_w$ | $R_w$ | $F1_w$ | $P_m$ | $R_m$ | $F1_m$ | Accuracy |
|---|---|---|---|---|---|---|---|
| DistilRoBERTa | 0.6933 | 0.6697 | 0.6593 | 0.6933 | 0.6697 | 0.6593 | 0.6697 |
| RoBERTa-base | 0.7026 | 0.6755 | 0.6625 | 0.7026 | 0.6755 | 0.6625 | 0.6755 |
| Robertuito | 0.6844 | 0.6575 | 0.6444 | 0.6844 | 0.6575 | 0.6444 | 0.6575 |
| BERT-base | 0.6795 | 0.6548 | 0.6378 | 0.6795 | 0.6548 | 0.6378 | 0.6548 |
| XLM-RoBERTa | 0.7256 | 0.7127 | 0.7011 | 0.7256 | 0.7127 | 0.7011 | 0.7127 |

Table 16: Transformer Model Results for Task 2 (Balanced Dataset)

| Model | $P_w$ | $R_w$ | $F1_w$ | $P_m$ | $R_m$ | $F1_m$ | Accuracy |
|---|---|---|---|---|---|---|---|
| DistilRoBERTa | 0.6425 | 0.6199 | 0.5934 | 0.6427 | 0.6202 | 0.5936 | 0.6202 |
| RoBERTa-base | 0.6296 | 0.6176 | 0.5901 | 0.6296 | 0.6176 | 0.5901 | 0.6179 |
| Robertuito | 0.5969 | 0.5758 | 0.5505 | 0.5969 | 0.5758 | 0.5505 | 0.5758 |
| BERT-base | 0.6125 | 0.5693 | 0.5441 | 0.6123 | 0.5691 | 0.5439 | 0.5693 |
| XLM-RoBERTa | 0.6099 | 0.5640 | 0.5370 | 0.6097 | 0.5638 | 0.5367 | 0.5640 |

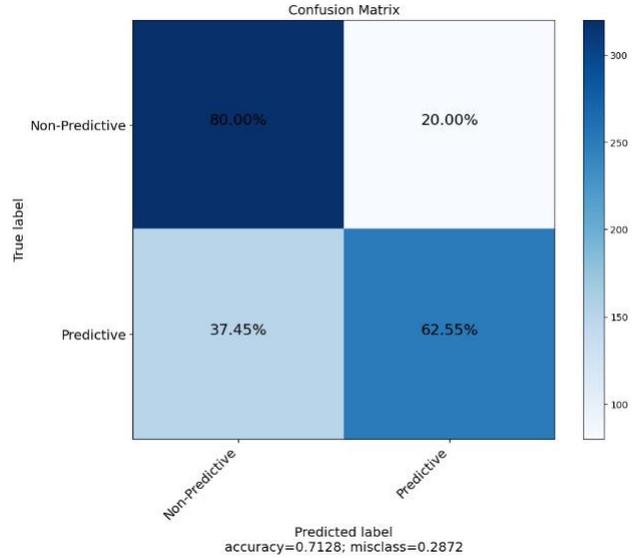

Figure 1: Confusion matrix illustrating classification performance between Non-Predictive and Predictive classes in Task 1.

## 8. Error analysis

Task 1 (Binary classification: Non-Predictive vs. Predictive): Both classes achieve reasonably balanced precision and recall, with Non-Predictive slightly outperforming Predictive. The F1-scores show moderate performance overall, indicating the model can distinguish the classes but with some room for improvement on the Predictive class.

Task 2 (Multiclass classification: Predictive Decremental, Incremental, Neutral): The Predictive Neutral class is classified with high precision and recall (around 82-84%), showing strong model confidence and accuracy. The two other classes have lower precision and recall ( 55-63%), suggesting the model struggles more with distinguishing incremental and decremental predictive statements. This is common in nuanced multiclass problems with overlapping features.

## 9. Discussion

This study highlights the importance of predictive statement classification in understanding the dynamics of investor sentiment in cryptocurrency-related social media content. By introducing a two-stage classification framework, we successfully distinguished between general predictive and non-predictive statements (Task 1) and further categorized predictive statements into Incremental, Decremental, and Neutral classes (Task 2).

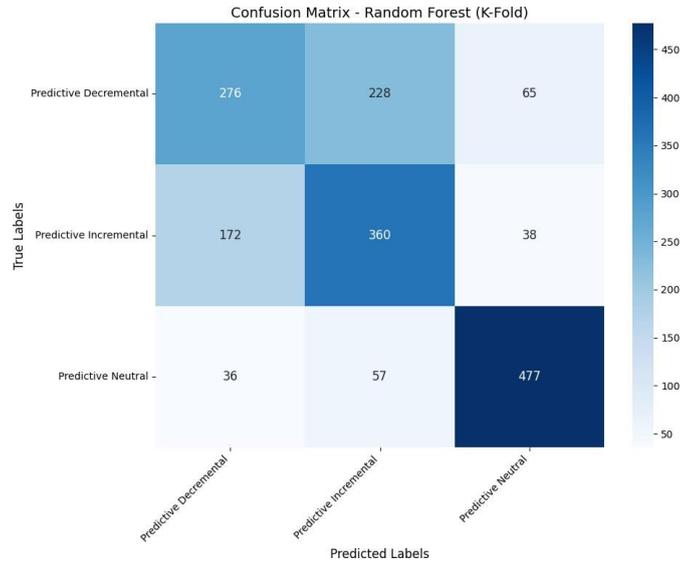

Figure 2: Confusion matrix showing Random Forest predictions across Decremental, Incremental, and Neutral classes in Task 2.

Table 17: Classification Report for Task 1 and Task 2

| Task / Label | Precision | Recall | F1-Score |
|---|---|---|---|
| **Task 1** | | | |
| Non-Predictive | 0.7334 | 0.8000 | 0.7524 |
| Predictive | 0.7178 | 0.6255 | 0.6498 |
| **Task 2** | | | |
| Predictive Decremental | 0.5702 | 0.4851 | 0.5242 |
| Predictive Incremental | 0.5581 | 0.6316 | 0.5926 |
| Predictive Neutral | 0.8224 | 0.8368 | 0.8296 |

The emotion analysis revealed clear distinctions in sentiment across prediction categories. Comments labeled as Incremental were strongly associated with positive emotions such as Delight and Joy and Enthusiasm and Eagerness, suggesting optimism and confidence in future price increases. In contrast, Decremental statements often co-occurred with negative emotions like Fear and Anxiety, Rage and Anger, and Grief and Sadness, reflecting user concerns or panic about anticipated downturns. However, a noteworthy finding is that some decremental predictions were still expressed with positive emotions. This indicates that certain users may view anticipated price drops as favorable opportunities—such as entry points for buying—which would not be captured through sentiment alone. This underscores the impact of our predictive framework: by separating sentiment from the type of prediction (incremental, decremental, or neutral), we enable more nuanced modeling of market psychology. Such granularity could significantly enhance price forecasting when emotional trends and predictive statements are analyzed in parallel with market movements over the same time intervals.

From a decision-making perspective, this separation is particularly valuable. Traditional sentiment analysis may treat all positive emotions as bullish and all negative emotions as bearish, which can lead to misleading interpretations. Our results show that the same price movement can generate opposite emotions depending on the investor's perspective (e.g., joy at price increases for holders versus frustration for those who missed the buying opportunity). By explicitly distinguishing Incremental, Decremental, and Neutral predictive statements, our framework provides clearer signals that investors can use to interpret market expectations, regulators can use to detect speculative bubbles or panic, and analysts can integrate with technical indicators to strengthen forecasting models. In this way, predictive statement classification directly contributes to more informed decision-making in highly volatile cryptocurrency markets.

When evaluating model performance, we observed significant improvements when training on the balanced dataset. For Task 1, the best F1-score (0.7011) was achieved by the transformer-based model XLM-RoBERTa, followed by SVM with RBF kernel (0.6825) among traditional models, and BiLSTM with GloVe embeddings (0.6349) as the top deep learning approach. In Task 2, where the classification challenge was finer-grained, Random Forest achieved the highest F1-score (0.6488) among traditional models, while CNN with FastText led the deep learning models (0.6153). DistilRoBERTa was the best-performing transformer with an F1-score of 0.5936. These results confirm that while transformer models generally outperform others, deep and traditional models remain competitive when class distributions are well-balanced.

## 10. Conclusion and Future Work

This study introduced a novel framework for classifying predictive statements in cryptocurrency-related tweets, distinguishing not only between predictive and non-predictive content but also among Incremental, Decremental, and Neutral predictions. By decoupling directional prediction from emotional tone, our approach captures the complexity of investor sentiment—particularly in cases where users express positive emotions even while anticipating a market decline. This insight underscores the need for more fine-grained sentiment analysis in financial contexts.

We employed GPT-based data augmentation to balance the dataset and used SenticNet to extract emotional signals. The results show that transformer-based models, especially XLM-RoBERTa, outperformed others in binary classification, while traditional models such as Random Forest showed superior results in multi-class classification. Deep learning approaches like BiLSTM and CNN also proved effective, highlighting that multiple model types can excel depending on task complexity and data structure.

Looking forward, we plan to align predictive statement types and emotional signals with actual price movements to assess their forecasting potential. Future work will also explore multilingual extensions, real-time applications, and alternative methods for dataset balancing—for example, approaches like PBC4cip (Loyola-González et al., 2017) to broaden the framework's impact in financial prediction and market behavior analysis.

## 11. Limitation

While our framework effectively distinguishes predictive statement types and associated emotions, it relies solely on textual data from a single platform (Twitter/X) and English-language content. This limits generalizability across languages, platforms, and cultural contexts. Additionally, our approach does not yet integrate real-time market data, which would be essential for validating the predictive power of the classified statements. Future work should address these aspects to enhance applicability and robustness.

**Acknowledgements**

The work was done with partial support from the Mexican Government through the grant A1-S-47854 of CONACYT, Mexico, grants 20241816, 20241819, and 20240951 of the Secretaría de Investigación y Posgrado of the Instituto Politécnico Nacional, Mexico. The authors thank the CONACYT for the computing resources brought to them through the Plataforma de Aprendizaje Profundo para Tecnologías del Lenguaje of the Laboratorio de Supercómputo of the INAOE, Mexico and acknowledge the support of Microsoft through the Microsoft Latin America PhD Award.## References

Ahani, Z., Tash, M.S., Balouchzahi, F., Ramos, L., Sidorov, G., Gelbukh, A., 2024. Social support detection from social media texts. arXiv preprint arXiv:2411.02580 .

Derczynski, L., 2016. Complementarity, f-score, and nlp evaluation, in: Proceedings of the Tenth International Conference on Language Resources and Evaluation (LREC'16), pp. 261–266.

Dulău, T.M., Dulău, M., 2019. Cryptocurrency–sentiment analysis in social media. Acta Marisiensis. Seria Technologica 16, 1–6.

Golnari, A., Komeili, M.H., Azizi, Z., 2024. Probabilistic deep learning and transfer learning for robust cryptocurrency price prediction. Expert Systems with Applications 255, 124404.

Islam, M.S., Bashir, M., Rahman, S., Al Montaser, M.A., Bortty, J., Nishan, A., Haque, M.R., 2025. Machine learning-based cryptocurrency prediction: Enhancing market forecasting with advanced predictive models. Journal of Ecohumanism 4, 2498–2519.

Kehinde, T., Adedokun, O.J., Joseph, A., Kabirat, K.M., Akano, H.A., Olanrewaju, O.A., 2025. Helformer: an attention-based deep learning model for cryptocurrency price forecasting. Journal of Big Data 12, 81.

Kolesnikova, O., Tash, M.S., Ahani, Z., Agrawal, A., Monroy, R., Sidorov, G., 2025. Advanced machine learning techniques for social support detection on social media. arXiv preprint arXiv:2501.03370 .

Kumar, D., Christopher, D., Balamurali, A., et al., 2024. Stock and cryptocurrency price prediction using svm and lstm algorithm, in: 2024 4th Asian Conference on Innovation in Technology (ASIANCON), IEEE. pp. 1–5.

Loyola-González, O., Medina-Pérez, M.A., Martínez-Trinidad, J.F., Carrasco-Ochoa, J.A., Monroy, R., García-Borroto, M., 2017. Pbc4cip: A new contrast pattern-based classifier for class imbalance problems. Knowledge-Based Systems 115, 100–109.

Nakamoto, S., 2008. Bitcoin: A peer-to-peer electronic cash system .

Otabek, S., Choi, J., 2024. From prediction to profit: A comprehensive review of cryptocurrency trading strategies and price forecasting techniques. IEEE Access .

Papacharalampous, G., Tyralis, H., 2022. A review of machine learning concepts and methods for addressing challenges in probabilistic hydrological post-processing and forecasting. Frontiers in Water 4, 961954.

Pellicani, A., Pio, G., Ceci, M., 2025. Carrot: Simultaneous prediction of anomalies from groups of correlated cryptocurrency trends. Expert Systems with Applications 260, 125457.

Qureshi, S.M., Saeed, A., Ahmad, F., Khattak, A.R., Almotiri, S.H., Al Ghamdi, M.A., Rukh, M.S., 2025. Evaluating machine learning models for predictive accuracy in cryptocurrency price forecasting. PeerJ Computer Science 11, e2626.

Roelleke, T., Wang, J., 2008. Tf-idf uncovered: a study of theories and probabilities, in: Proceedings of the 31st annual international ACM SIGIR conference on Research and development in information retrieval, pp. 435–442.

SenticNet, . Sentic api. https://sentic.net/api/. Accessed: 2024-07-02.

Shahiki Tash, M., Ahani, Z., Tash, M., Kolesnikova, O., Sidorov, G., 2024. Analyzing emotional trends from x platform using senticnet: A comparative analysis with cryptocurrency price. Cognitive Computation , 1–18.

Sizan, M.M.H., Das, B.C., Shawon, R.E.R., Rana, M.S., Al Montaser, M.A., Chouksey, A., Pant, L., 2023. Ai-enhanced stock market prediction: Evaluating machine learning models for financial forecasting in the usa. Journal of Business and Management Studies 5, 152–166.

Tash, M.S., Ahani, Z., Kolesnikova, O., Sidorov, G., 2024a. Analyzing emotional trends from x platform using senticnet: A comparative analysis with cryptocurrency price. arXiv preprint arXiv:2405.03084 .

Tash, M.S., Kolesnikova, O., Ahani, Z., Sidorov, G., 2024b. Psycholinguistic and emotion analysis of cryptocurrency discourse on x platform. Scientific Reports 14, 8585.

Viéitez, A., Santos, M., Naranjo, R., 2024. Machine learning ethereum cryptocurrency prediction and knowledge-based investment strategies. Knowledge-Based Systems 299, 112088.